\documentclass[12pt]{article}

\usepackage{authblk}

\usepackage{listings} 
\usepackage{xcolor}
 
\definecolor{codegreen}{rgb}{0,0.6,0}
\definecolor{codegray}{rgb}{0.5,0.5,0.5}
\definecolor{codepurple}{rgb}{0.58,0,0.82}
\definecolor{backcolour}{rgb}{0.95,0.95,0.92}
 
\lstdefinestyle{mystyle}{
    backgroundcolor=\color{backcolour},   
    commentstyle=\color{codegreen},
    keywordstyle=\color{magenta},
    numberstyle=\tiny\color{codegray},
    stringstyle=\color{codepurple},
    basicstyle=\ttfamily\footnotesize,  
    breakatwhitespace=false,         
    breaklines=true,                 
    captionpos=b,                    
    keepspaces=true,                 
    numbers=left,                    
    numbersep=5pt,                  
    showspaces=false,                
    showstringspaces=false,
    showtabs=false,                  
    tabsize=2
}
\usepackage[utf8]{inputenc}
\usepackage[english]{babel}

\usepackage{graphicx}
\graphicspath{ {./images/} }
\usepackage[export]{adjustbox}
\usepackage{subcaption}

\usepackage{pgfplotstable}
\usepackage{siunitx}
\usepackage{booktabs} 

\usepackage{caption}

\usepackage{framed}
\usepackage{authblk}

\usepackage{enumitem}
\newlist{todolist}{itemize}{2}
\setlist[todolist]{label=$\square$}

\usepackage{amssymb}
\usepackage{amsmath, bm}
\usepackage{xifthen}
\newcommand{\E}[2][]{%
  \ifthenelse{\isempty{#1}}{%
    \mathrm{E}\left[#2\right]
  }{%
    \mathrm{E}\left[#2 \middle| #1\right]
  }%
}

\newcommand{\Cov}[2][]{%
  \ifthenelse{\isempty{#1}}{%
    \mathrm{Cov}\left(#2\right)
  }{%
    \mathrm{Cov}\left(#2, #1\right)
  }%
}

\usepackage{verbatim}

\usepackage{natbib}
\newcommand{\blind}{0}

\def\spacingset#1{\renewcommand{\baselinestretch}%
{#1}\small\normalsize} \spacingset{1}

\title{You Only Compress Once: Optimal Data Compression for Estimating Linear Models}

\begin{document}

\if0\blind
{
  \author[1]{Jeffrey Wong}
  \author[1]{Eskil Forsell}
  \author[2]{Randall Lewis\thanks{Lewis was employed at Netflix when this work began.}}
  \author[1]{Tobias Mao}
  \author[1]{Matthew Wardrop}
  \affil[1]{Netflix, Inc.}
  \affil[2]{Nanigans, Inc.}
  \maketitle
} \fi

\if1\blind
{
  \bigskip
  \bigskip
  \bigskip
  \begin{center}
    {\LARGE\bf You Only Compress Once: Optimal Data Compression for Estimating Linear Models}
  \end{center}
  \medskip
} \fi

\begin{abstract}
    Linear models are used in online decision making, such as in machine learning, policy algorithms, and experimentation platforms. Many engineering systems that use linear models achieve computational efficiency through distributed systems and expert configuration. 
    While there are strengths to this approach, it is still difficult to have an environment that enables researchers to interactively
    iterate and explore data and models, as well as leverage analytics solutions from the open source community. Consequently, innovation can be blocked. 
    
    Conditionally sufficient statistics is a unified data compression and estimation strategy that is useful for the model development process, as well as the engineering deployment process. The strategy estimates linear models from compressed data without loss on the estimated parameters and their covariances, even when errors are autocorrelated within clusters of observations. Additionally, the compression preserves almost all interactions with the the original data, unlocking better productivity for both researchers and engineering systems.
\end{abstract}

\noindent
{\it Keywords:} Algorithms, Statistical Computing, Machine Learning, Experimentation, Econometrics.

\spacingset{1.5}
\section{Introduction}


Linear models are highly versatile and are commonly used in machine learning and causal inference. Applications in the former field include multi-armed bandit problems for algorithmic decision making \citep{agrawal2013thompson}, and in the latter, estimating average treatment effects, conditional average effects, and time-dynamic effects, and improving statistical power. Modern experimentation platforms (XPs), the main focus of this paper, often aim to enable methods from both fields. However,
implementing linear models in a large and interactive engineering system has several challenges.
First, it must be able to scale both to large sample sizes, which can be as large as hundreds of millions of observations, and to many features, sometimes in the thousands. Second, it should be reproducible and extensible such that software engineers and researchers can interact with, iterate on, and subsequently contribute to it \citep{diamantopoulos2020engineering}.


Regarding the first challenge, XP communities have found solutions to realize most of the gains from linear models, such as statistical power, while still having a highly scalable solution. CUPED \citep{deng2013improving} is such an implementation that has been adopted by at least Microsoft, Uber \citep{uber}, and Booking.com \citep{booking}. Because these implementations run on distributed systems, they introduce network latency and require expert maintenance and configuration, making them difficult to reproduce, interact with, and extend.
Despite the ability to meet the demands of large sample sizes and large feature sets, engagement and contributions from the community can be limited due to a high level of expertise needed to extend the online environment. This creates a divide between what is feasible in offline model development, where barriers are lower, and what is feasible in online deployment.

Addressing the second challenge, Netflix described an inclusive XP that makes use of single-machine computation for modeling, allowing it to be more interactive and consistent with the way researchers iterate \citep{diamantopoulos2020engineering}. 
As a result, researchers can reproduce analyses from the XP, iterate, follow up, and debug using Python and R, and then contribute improvements to statistical methodology back to the engineering systems.
Netflix called this a ``technical symbiosis" that can make what is feasible in offline model development become feasible in online deployment, ultimately leading to many success stories for the business in \citet{forsell2020success}.

We further these ideas by offering a compression and estimation strategy for large linear models that improves performance, while maintaining the ability to explore data interactively. Conditionally sufficient statistics, described in Section 4, reduces data volume,
allowing researchers to operate on a small data frame where they can explore the data interactively, just as they would with uncompressed data. It also allows lossless estimation of ordinary least squares (OLS) with homoskedastic, heteroskedastic, and clustered covariances, and similarly for other generalized linear models. Multiple outcome variables can be estimated from a common data structure, making the analysis of multiple metrics easy.

This compression strategy is a significant deviation from other literature that discusses distributed computing, parallelization, or SGD, which can reduce computing time but do not resolve challenges with data volume. Our contribution is unique because it reduces both computing time and data volume, and can also be combined with the above strategies.
Several linear models have become tractable with single-machine computation, even on datasets with a sample size of 50 million. Such an efficient and interactive computing environment opens the modeling backend of an XP to implementations from Python and R, whose libraries have historically focused on single-machine implementations. Having this environment also reduces the differences between offline model development and online deployment, increasing agility and productivity.

\section{Setting}
For the remainder of the paper, consider the setting in which there are $n$ observations consisting of vectors $\begin{pmatrix}\mathbf{y}_i^\top & \mathbf{m}_i^\top\end{pmatrix}$ where $\mathbf{y}_i$ is a length $o$ column vector of outcomes, and $\mathbf{m}_i$ is a length $p$ column vector of covariates. These observations are stacked into the outcome matrix, $\mathbf{y}\in\mathbb{R}^{n \times o}$, and the feature matrix, $\mathbf{M} \in \mathbb{R}^{n \times p}$. For the remainder of the text we focus on the case where $o=1$ and the outcome matrix $\mathbf{y}$ is simply a column vector, but the results trivially extend to the $o > 1$ case. 
We wish to estimate the linear model,
\begin{align*}
    \mathbf{y} = \mathbf{M} \bm{\beta} + \bm{\varepsilon},
\end{align*}
where the first two moments of $\bm{\varepsilon}$ have the following structure:
\begin{align*}
    \mathbb{E}[\bm{\varepsilon}|\mathbf{M}]&=0\text{, and}\\ \mathbb{E}[\bm{\varepsilon}\bm{\varepsilon}^\top|\mathbf{M}]&=\mathbf{\Omega}.
\end{align*}
Using ordinary least squares (OLS), the estimate of $\bm{\beta}$ and its covariance are
    \begin{align*}
        \hat{\bm{\beta}} &= (\mathbf{M}^\top \mathbf{M})^{-1} (\mathbf{M}^\top \mathbf{y}) \text{ and}\\
        \mathbb{V}(\hat{\bm{\beta}}) &= (\mathbf{M}^\top \mathbf{M})^{-1} (\mathbf{M}^\top \Omega \mathbf{M}) (\mathbf{M}^\top \mathbf{M})^{-1}.
    \end{align*}

This expression for $\mathbb{V}(\hat{\bm{\beta}})$ is known as the sandwich covariance matrix \citep{huber1967heteroskedasticity}, which is the basis for estimators for $\mathbb{V}(\hat{\bm{\beta}})$ under different structures of $\mathbf{\Omega}$. It has gained its name due to its similarities to ``meat'', $\mathbf{\Xi}=\mathbf{M}^\top \Omega \mathbf{M}$, placed between two pieces of ``bread'', $\mathbf{\Pi}=(\mathbf{M}^\top \mathbf{M})^{-1}$. The primary contribution of this paper is showing compression strategies for $\mathbf{M}$ that work under the three most common structures of $\mathbf{\Omega}$.

To make the connection to databases or dataframes, we call a row in such a structure a record which consists of a single observation. However, we will also discuss a compressed record, a row which represents multiple observations by including a weight denoting the number of observations the compressed record represents.

\section{Previous Work}
There have been previous attempts to circumvent the need to have access to the full data when estimating linear models. In this section we will outline the four most common strategies and discuss how they relate to the needs of an XP.

\subsection{T-tests}
Given two randomized and controlled samples, one representing the treatment group and the other the control group, a standard two-sample t-test can be estimated from aggregates, the means and variances of each sample. Alternatively, a t-test can also operate on unaggregated data, and is equivalent to estimating an OLS model with an intercept and an indicator for treatment, as shown in \citet{wong2019efficient}. This relationship may suggest that it is possible to estimate OLS models with more parameters using data that is already aggregated; we will show that this is indeed the case below.

\subsection{Streaming Algorithms}
Streaming algorithms, such as Stochastic Gradient Descent (SGD) \citep{bottou2010large}, also circumvent the need of having access to the full dataset. In contrast to the direct algebraic solution above, these computing methods do not need the data to fit into memory at one time; instead, they read data from disk in batches in order to update the estimate of $\bm{\beta}$. SGD is a highly specialized solution for estimating models on large volumes of data, even when using a single machine, and is built into machine learning software such as Vowpal Wabbit \citep{vowpal}. However, reading from disk causes a decrease in performance, and without a holistic streaming solution for other statistics and visualizations, it is still difficult to explore large volumes of data. The method we present below compresses data to enable the algebraic solution, but SGD and its variants can also operate on compressed data, making our contribution complementary to this line of work.

\subsection{Frequency Weights}
Another literature proposes the use of frequency weights (f-weights), for example in SAS \citep{fweight}.
The compression strategy is simple: count and collapse identical observations into one compressed record and assign an f-weight equal to the number of duplicate observations. This compression is lossless: even though we record one single compressed record in the dataset, we can still recover the original uncompressed observations. Statistical functions that are compatible with f-weights are available in software such as SAS and Stata. 

Because the compression is lossless we can estimate the distribution of $\hat{\bm{\beta}}$. Let $(\dot{\mathbf{y}}, \dot{\mathbf{M}})$ be the compressed data, and let $\dot{\mathbf{n}}$ be the vector of f-weights. Then, using weighted OLS (WLS),
\begin{align*}
    \hat{\bm{\beta}} &= (\dot{\mathbf{M}}^\top \mathbf{W} \dot{\mathbf{M}})^{-1} (\dot{\mathbf{M}}^\top \mathbf{W} \dot{\mathbf{y}}), \text{and} \\
    \mathbb{V}(\hat{\bm{\beta}}) &= 
    (\dot{\mathbf{M}}^\top \mathbf{W} \dot{\mathbf{M}})^{-1}
    (\dot{\mathbf{M}}^\top  \sqrt{\mathbf{W}} \dot{\mathbf{\Omega}} \sqrt{\mathbf{W}} \dot{\mathbf{M}})
    (\dot{\mathbf{M}}^\top \mathbf{W} \dot{\mathbf{M}})^{-1}
\end{align*}
where $\mathbf{W}$ is a diagonal matrix with $\dot{\mathbf{n}}$ on the diagonal and $\dot{\mathbf{\Omega}}$ is $\mathbf{\Omega}$ after deduplication for each compressed record. 
Unfortunately this method relies on having duplicate observations in $(\dot{\mathbf{y}}, \dot{\mathbf{M}})$ which is unlikely except in special cases.

\subsection{Group Regression}
\label{groupregression}
Group regression \citep{angrist2008mostly, chamberlain1984handbook} is used in applications where the researcher lacks access to the individual-level records but does have access to group-level aggregates, such as economic and demographic data by state or county. In such settings, the model coefficients, $\hat{\bm{\beta}}$, can still be losslessly recovered from the weighted regression
\begin{align*}
    \bar{\mathbf{y}} = \bar{\mathbf{M}} \hat{\bm{\beta}}
\end{align*}
with group sizes $\bar{\mathbf{n}}$ as weights. Here, $\bar{\mathbf{y}}$ is a column vector of group means and $\bar{\mathbf{M}}$ is a group-level feature matrix. This method only requires the group mean which is usually directly recorded or computable from group aggregates such as the group sum and size. However, estimates of $\mathbb{V}(\hat{\bm{\beta}})$ are noisier due to the absence of a sufficient statistic: the variance for each group. This is especially problematic when a compressed record summarizes multiple individual records, but is also the setting in which there is most benefit to computation. The method we propose removes that conflict by defining the sufficient statistics that must be recorded at compression time in order to losslessly recover $\mathbb{V}(\hat{\bm{\beta}})$.

\section{Lossless Compression with Sufficient Statistics \label{sec:lossless_compression_with_suff_stats}}

Generalized linear models (GLMs) are based on the family of exponential distributions, which have a natural compression strategy. These distributions have unknown parameters, $\bm{\theta}$, which are estimated either from a sample, $\mathbf{y}$, or from a set of aggregates known as the sufficient statistics, $T(\mathbf{y})$ \citep{lehmann2006theory}. For example, when samples are drawn independently, the mean and variance parameters of a Gaussian distribution can be estimated using the aggregates $T(\mathbf{y}) = \{\sum\limits_i y_i=y', \sum\limits_i y_i^2=y'', n\}$, where $n$ is the sample size.

GLMs model the parameters of an exponential distribution that condition on a vector of features. We extend the concept of sufficient statistics to conditionally sufficient statistics. Given a feature matrix $\mathbf{M}$, and a feature vector $\mathbf{m}^*$, $T(\mathbf{y} | \mathbf{m}^*) = \{\sum\limits_{i | \mathbf{m}_i = \mathbf{m}^*} y_i,\; \allowbreak
  \sum\limits_{i | \mathbf{m}_i = \mathbf{m}^*} y_i^2,\; \allowbreak
  \sum\limits_{i | \mathbf{m}_i = \mathbf{m}^*} 1\}$ are conditionally sufficient statistics.
From $n$ data points, a linear model can be learned by first stacking unique feature vectors $\mathbf{\tilde{m}}_1^\top \ldots \mathbf{\tilde{m}}_G^\top$ into a feature matrix $\tilde{\mathbf{M}}$, and then stacking the sum, sum of squares, and counts for each distinct response vector into column vectors $\tilde{\mathbf{y}}'$, $\tilde{\mathbf{y}}''$, and $\tilde{\mathbf{n}}$. 
When data is stored in a database, we can accomplish this by grouping by the features and computing three aggregates.
We can then estimate the weighted linear model:
\begin{equation}
\frac{\tilde{\mathbf{y}}'}{\tilde{\mathbf{n}}} = \tilde{\mathbf{M}} \bm{\beta} + \bm{\varepsilon}
\label{eq:group_regression}
\end{equation}
with weights $\tilde{\mathbf{n}}$, and where $\frac{\tilde{\mathbf{y}}'}{\tilde{\mathbf{n}}}$ uses element-wise division. This compressed regression is equivalent to group regression and operates on $G$ compressed records instead of $n$ or $\dot{n}$, as illustrated with an example in Table~\ref{tab:compression}. Weighted least squares (WLS) coefficient estimates of this compressed model are mathematically equivalent and numerically identical to the OLS coefficients of the uncompressed model. The computational complexity for estimating least squares is linear in the number of compressed records, so the time to estimate the model can be significantly reduced using conditionally sufficient statistics.

\sisetup{                           
        round-mode      = places,   
        }

\begin{table}
    \centering
    \caption{Example dataset and its compressed versions.}
    \begin{subtable}[t]{.14\linewidth}
        \centering 
        \caption{}
        \pgfplotstableread[col sep=comma]{tables/compression_data_a_full.csv}\fulldata
        \pgfplotstabletypeset[
        every head row/.style={before row=\toprule,after row=\midrule},
        every last row/.style={after row={
			\bottomrule}
			},
        display columns/0/.style={string type,column name={$\mathbf{M}$}},
        display columns/1/.style={column name={$\mathbf{y}$}, precision=3,column type={S[]}, column type = {r}},
        multicolumn names
        ]
        {\fulldata}
\vspace{2mm}
    \end{subtable}
    \begin{subtable}[t]{.28\linewidth}
        \centering
        \caption{}
        \pgfplotstableread[col sep=comma]{tables/compression_data_b_fweight.csv}\fweightdata
        \pgfplotstabletypeset[
        every head row/.style={before row=\toprule,after row=\midrule},
        every last row/.style={after row={\bottomrule}},
        display columns/0/.style={string type,column name={$\dot{\mathbf{M}}$}},
        display columns/1/.style={column name={\multicolumn{0}{c}{$\dot{\mathbf{y}}$}}, precision=2,column type={S[]},column type={r}},
        display columns/2/.style={column name={\multicolumn{0}{c}{$\dot{\mathbf{n}}$}},column type={S[]}, column type={r}}
        ]
        {\fweightdata}
    \end{subtable}
    \begin{subtable}[t]{.26\linewidth}
        \centering
        \caption{}
        \pgfplotstableread[col sep=comma]{tables/compression_data_c_group.csv}\groupdata
        \pgfplotstabletypeset[
        every head row/.style={before row=\toprule,after row=\midrule},
        every last row/.style={after row={\bottomrule}},
        display columns/0/.style={string type,column name={$\bar{\mathbf{M}}$}},
        display columns/1/.style={column name={\multicolumn{0}{c}{$\bar{\mathbf{y}}$}}, precision=2,column type={S[]},column type={r}},
        display columns/2/.style={column name={\multicolumn{0}{c}{$\bar{\mathbf{n}}$}},column type={r}}
        ]
        {\groupdata}
    \end{subtable}
    \begin{subtable}[t]{.26\linewidth}
        \centering
        \caption{}
        \pgfplotstableread[col sep=comma]{tables/compression_data_d_compressed.csv}\compresseddata
        \pgfplotstabletypeset[
        every head row/.style={before row=\toprule,after row=\midrule},
        every last row/.style={after row={\bottomrule}},
        display columns/0/.style={string type,column name={$\tilde{\mathbf{M}}$}},
        display columns/1/.style={column name={\multicolumn{0}{c}{$\tilde{\mathbf{y}}'$}}, precision=2,column type={S[]},column type={r}},
        display columns/2/.style={column name={\multicolumn{0}{c}{$\tilde{\mathbf{y}}''$}}, precision=2,column type={S[]},column type={r}},
        display columns/3/.style={column name={\multicolumn{0}{c}{$\tilde{\mathbf{n}}$}},column type={r}}
        ]
        {\compresseddata}
        
    \end{subtable}
    \captionsetup{justification=centering}
    \caption*{
        (a) Uncompressed data. (b) f-weights: $(\mathbf{y}, \mathbf{M})$-compressed records.\protect\\
        (c) Groups: $(\mathbf{M})$-compressed records. (d) Sufficient Statistics: $(\mathbf{M})$-compressed records.
    }
    \label{tab:compression}
\end{table}

While the point estimates are numerically identical, covariances from WLS on group means are not, as we saw in Section \ref{groupregression}. The sufficient statistics do, however, contain all the information needed to calculate covariances that are identical to those from the uncompressed model. In Section \ref{Sandwich} we describe how to do this for three different structures of the covariance matrix.

In Table~\ref{tab:compression} we provide examples of all compression strategies, then in Table~\ref{tab:compression_comparison} we review the trade-offs of each strategy. The compression strategies trade-off conceptual simplicity in exchange for computational efficiency. For example, ``uncompressed'' provides lossless analysis using OLS. F-weights provide some compression, but need a separate compression for each outcome. Groups provide greater compression by aggregating over each outcome, granting what we call the ``You Only Compress Once'' (YOCO) property, but come at the cost of a lossy variance estimator. Finally, sufficient statistics power the complete recovery of the uncompressed distribution by precomputing the sufficient statistics and adjusting the formulas for $\mathbb{V}(\hat{\bm{\beta}})$. As a result, the experimentation platform can realize the computational performance gains without any loss in quality of the final results.

\begin{table}[b]
 \small
    \centering
    \caption{Comparison of Compression Strategies}
    \begin{tabular}{l|ccccc}
        \phantom{(a)} Strategy & Compression & Record & Estimator & $\mathbb{V}(\hat{\bm{\beta}})$ & YOCO($\mathbf{y_i}$)* \\
        \hline
        (a) Uncompressed & - & $(\mathbf{m_i}; y_i)^\top$ & OLS & Lossless & - \\
        (b) f-weights & Good: $(\mathbf{y}, \mathbf{M})$ & $(\dot{\mathbf{m}}\mathbf{_g}; \dot{\mathbf{y}}\mathbf{_g'}; \dot{\mathbf{n}}\mathbf{_g})^\top$ & WLS & Lossless & No \\
        (c) Groups & Best: $(\mathbf{M})$ & $(\mathbf{\tilde{m}_g}; \mathbf{\tilde{y}_g'}; \mathbf{\tilde{n}_g})^\top$ & WLS & \textit{Lossy} & Yes \\ 
        (d) Sufficient Statistics & Best: $(\mathbf{M})$ & $(\mathbf{\tilde{m}_g}; \mathbf{\tilde{y}_g'}; \mathbf{\tilde{y}_g''}; \mathbf{\tilde{n}_g})^\top$ & WLS & Lossless & Yes \\
        \multicolumn{6}{l}{\footnotesize * ``You Only Compress Once'' (YOCO) across multiple outcomes $\mathbf{y_i}$ without losing compression.}
    \end{tabular}
    \label{tab:compression_comparison}
\end{table}

\subsection{Interactivity}

Even when the original dataset is discarded and only the compressed datasets $\mathbf{\tilde{M}}$, $\tilde{\mathbf{y}}'$, $\tilde{\mathbf{y}}''$ and $\tilde{\mathbf{n}}$ are retained,
it is still possible for a researcher to do exploratory data analysis. For example, using $\mathbf{\tilde{M}}$ and $\mathbf{\tilde{n}}$ we can compute summary statistics on the features using weighted means, medians, or quantiles. It is also possible to examine the correlation or co-occurrence between two features. The histogram of the features can be plotted. The mean and variance of $\mathbf{y}$ can be estimated, and the relationship between the expected value of $\mathbf{y}$ and a feature in $\mathbf{\tilde{M}}$ can also be plotted. In the extreme, new features based on $\mathbf{\tilde{M}}$ can be generated and added to the linear model, for example an interaction feature. This interactivity and fast linear modeling is a powerful combination that can minimize context switches and accelerate research cycles.

\section{The Sandwich Covariance Matrix}
\label{Sandwich}

In this section we will outline compression strategies under three common structures of $\mathbf{\Omega}$: (1) homoskedastic covariances, where $\mathbf{\Omega}$ is a diagonal matrix with a constant on the diagonal; (2) heteroskedastic covariances, where $\mathbf{\Omega}$ is a diagonal matrix but its entries are a function of the features; and (3) cluster robust covariances, where $\mathbf{\Omega}$ is a block diagonal matrix and its entries are also a function of the features.

The ``bread'' of the sandwich can be computed from compressed records as
$$\mathbf{\Pi}=(\mathbf{M}^\top \mathbf{M})^{-1} = (\tilde{\mathbf{M}}^\top \text{diag}(\tilde{\mathbf{n}}) \tilde{\mathbf{M}})^{-1}.$$
As this is independent of $\mathbf{\Omega}$ we will only discuss compression strategies for computing the ``meat'' matrix, $\mathbf{\Xi}=\mathbf{M}^\top \Omega \mathbf{M}$, below.

\subsection{Homoskedastic Covariances}

In the textbook OLS case where errors are assumed to be i.i.d., $\mathbf{\Omega} = \sigma^2 I_n$ and thus homoskedastic, which leads to 
\begin{align*}
    \mathbf{\Xi}_\text{OLS} &=\sigma^2 \mathbf{M}^\top \mathbf{M}\\
    &=\sigma^2\mathbf{\Pi}^{-1}.
\end{align*}
As $\mathbf{\Pi}$ is just the bread matrix we focus on estimating $\sigma^2$. Let $\hat{\mathbf{y}}=\mathbf{M}\hat{\bm{\beta}}$ be the fitted values and $\bm{e}=\mathbf{y}-\hat{\mathbf{y}}$ the residuals, the sample equivalent of $\bm{\varepsilon}$. The estimator for $\sigma^2$ can then be written as $\hat{\sigma}^2 = \frac{\sum\limits_i e_i^2}{n-p}$ where $\sum\limits_i e_i^2$ is commonly known as the residual sum of squares, RSS.

The RSS can be partitioned into $G$ sums representing the RSS for the $G$ unique feature vectors, $\mathbf{\tilde{m}}_1^\top \ldots \mathbf{\tilde{m}}_G^\top$. The compression of $\mathbf{M}$ to $\tilde{\mathbf{M}}$ creates groups where features are identical within a group, so observations in a group have the same fitted outcome, $\hat{y}_{g, i} = \hat{y}_{g}\, \forall i \in g$. We can define $\hat{\tilde{\mathbf{y}}}=\tilde{\mathbf{M}}\hat{\bm{\beta}}$ and reduce the RSS to

\begin{align*}
RSS &= \sum\limits_{g = 1}^{G} \sum\limits_{i = 1}^{\tilde{n}_g} (y_{g, i}-\hat{y}_{g,i})^2 \\
  &= \sum\limits_{g = 1}^{G} \left(\hat{\tilde{y}}_g^2 \tilde{n}_g - 2\hat{\tilde{y}}_{g} \sum\limits_{i = 1}^{\tilde{n}_g} y_{g,i} + \sum\limits_{i = 1}^{\tilde{n}_g} y_{g, i}^2\right) \\
  &= \sum\limits_{g = 1}^{G} \left(\hat{\tilde{y}}_g^2 \tilde{n}_g - 2\hat{\tilde{y}}_{g} \tilde{y}'_g + \tilde{y}''_g\right) = \sum\limits_{g = 1}^{G} \widetilde{RSS}_g,
\end{align*}
an operation that only requires the sufficient statistics but can fully recover $\hat{\sigma}^2$.

\subsection{Heteroskedasticity-Consistent Covariances}

Heteroskedasticity-consistent covariances are needed when errors are assumed to be i.i.d. only when conditioning on the features. This gives us the following structure of the covariance matrix and its standard Eicker-Huber-White (EHW) \citep{eicker1967heteroskedasticity,huber1967heteroskedasticity,white1980heteroskedasticity} estimator:
\begin{align*}
    \mathbf{\Xi}_\text{EHW} &=\mathbf{M}^\top \text{diag}(\bm{\sigma}^2) \mathbf{M},\\
    \hat{\mathbf{\Xi}}_\text{EHW} &= \mathbf{M}^\top\text{diag}\left(\mathbf{e}^2\right)\mathbf{M}\\
    &= \tilde{\mathbf{M}}^\top\text{diag}\left(\tilde{\mathbf{e}}''\right)\tilde{\mathbf{M}},
\end{align*}
where $\tilde{\mathbf{e}}''$ stacks the residual sum of squares for each group,
$\widetilde{RSS}_g$, described above.

It is common for XPs to analyze the impact on binary metrics. When using a linear probability model, heteroskedastic errors are guaranteed, motivating the use of these heteroskedasticity-consistent covariances.
Later, in Section 7, we will also show that logistic regression is compatible with our compression strategy.

\subsection{Cluster-Robust Covariances}

Cluster-robust covariances are needed for data that has autocorrelation within, but not between, clusters of observations. This structure is a core component of inference on panel data using pooled OLS or fixed effects \citep{cameron2015practitioner, woolridgePanel}. It is also a broad generalization of homoskedastic and heteroskedastic covariances. Throughout this section, we will rely on a motivating example of data with repeated observations, but all results directly extend to arbitrary feature matrices.

Suppose a study samples $n_u$ users randomly so that the users are independently and identically distributed. The users are then observed each day for $T$ days, with no loss to follow up, and a response variable, $y_{u,t}$, is measured. Some data about the users are known, summarized in feature matrix $\mathbf{M}_1$. 
For simplicity, say these covariates are measured prior to treatment, and are therefore constant during the $T$ days.
To complement these static covariates, let the time index, $\mathbf{t}=\{t_0, t_1, \ldots, T\}$, be a dynamic covariate we wish to use in the model and stack this in feature matrix $\mathbf{M}_2$. The full feature matrix is thus $\mathbf{M}=\begin{bmatrix}\mathbf{M}_1 & \mathbf{M}_2\end{bmatrix}$ and contains $n = n_u \cdot T$ records. This dataset can be used to estimate the model
$$\mathbf{y}_{u, t} = \alpha + \mathbf{M_1}\beta_1 + \mathbf{M_2}\beta_2 + \bm{\varepsilon}_{u,t},$$
which can be used to estimate a treatment effect while controlling for temporal variation. An interesting extension is $$\mathbf{y}_{u, t} = \alpha + \mathbf{M_1}\beta_1 + \mathbf{M_2}\beta_2 + \mathbf{M_3}\beta_3 + \bm{\varepsilon}_{u,t},$$
where $\mathbf{M}_3$ is the interaction of $\mathbf{M}_1$ and $\mathbf{M}_2$, allowing the researcher to estimate a treatment effect with time heterogeneity. This can be used to see how treatment effects saturate or diminish over time.
For example, \citet{fitzmaurice2008primer} discusses a similar model to analyze forced expiratory volume (FEV), a measurement of lung health.

The dataset with feature matrix $\mathbf{M}$ is a repeated observations dataset because there are multiple observations per user. Furthermore, there is autocorrelation within a user across time, but there is independence across users. In this example, the data is clustered by users, and the number of clusters is $C = n_u$.
Due to independence across users, and autocorrelation within users, the covariance matrix has the structure
$$
\mathbf{\Omega}=
    \begin{bmatrix}
        \mathbf{\Omega}_1 & & & & 0 \\
        & \ddots & & & \\
        & & \mathbf{\Omega}_c & & \\
        & & & \ddots & \\
        0 & & & & \mathbf{\Omega}_{C}
    \end{bmatrix}
$$
where $c$ is the cluster index, and $\mathbf{\Omega}_c$ is the covariance matrix for the observations within the cluster \citep{newey1986simple, liang1986cluster}. As this is a block-diagonal matrix
\begin{align*}
    \mathbf{\Xi}_\text{NW}&=\mathbf{M}^\top \text{diag}(\bm{\varepsilon}) \mathbf{W}_C \mathbf{W}_C^\top \text{diag}(\bm{\varepsilon}) \mathbf{M}\text{, and}\\
    \hat{\mathbf{\Xi}}_\text{NW} &= \mathbf{M}^\top \text{diag}(\bm{e}) \mathbf{W}_C \mathbf{W}_C^\top \text{diag}(\bm{e}) \mathbf{M} \\
    &= \textstyle\sum\limits_c\mathbf{M}_c^\top \bm{e}_c \bm{e}_c^\top \mathbf{M}_c
 \end{align*}
where $\mathbf{M}_c$ and $\mathbf{e}_c$ are the subsets of $\mathbf{M}$ and $\mathbf{e}$ for records belonging to cluster $c$, and $\mathbf{W}_C\in\mathbb{R}^{n\times C}$ is the cluster matrix with the entry in row $i$, column $c$, equal to $1$ if observation $i$ belongs to cluster $c$, and 0 otherwise. It is assumed that a single observation can only belong to one cluster.
Homoskedastic and heteroskedasticity-consistent covariances are special cases of cluster robust covariances where $n = C$. 

Estimating the regression coefficients and covariances for a repeated observations dataset can be challenging to do on a single machine. Suppose $n_u = C = 1 \cdot 10^7$, $T = 100$, and there are $p = 10$ covariates measured per user. If the data is stored with floating point precision, then 40 GB of memory are needed to hold the dataset. This is a large task even for a modern desktop computer, and may force a researcher to use a larger, remote server to analyze the data, or to use out of core computing methods. In contrast, the dataset without repeated observations only requires 400 MB. Computation on such a large dataset is expensive, and when forced into environments where the data cannot be stored in-memory, computation becomes even more costly. 
Below we outline three variations of the compression strategy to reduce data volume and computing cost while still estimating clustered covariances without loss. The efficiency of each of these methods is dependent on how the data is structured.

\subsubsection{Within-cluster Compression}
\label{withinclustercompression}

To calculate the contribution to the meat matrix for a given cluster, $c$, any compression strategy must retain some structure on the relationship between compressed records and clusters. In the simplest approach, each compressed record only contains data from a single cluster. This is trivially satisfied if the cluster identifier is completely determined by an observation's feature vector. It can also be achieved by adding the cluster identifier to the feature matrix,
compressing as in Section~\ref{sec:lossless_compression_with_suff_stats}, then discarding it after. In both cases $\tilde{\mathbf{M}}$ will contain $G\geq C$ compressed records.

The meat matrix can now be expressed as:
\begin{align*}
\hat{\mathbf{\Xi}} = \tilde{\mathbf{M}}^\top \text{diag}(\tilde{\bm{e}}') \tilde{\mathbf{W}}_C \tilde{\mathbf{W}}_C^\top \text{diag}(\tilde{\bm{e}}') \tilde{\mathbf{M}},
\end{align*}
where $\tilde{\mathbf{W}}_C\in\mathbb{R}^{G\times C}$ is the grouped cluster matrix with the entry in row $g$, column $c$ equal to $1$ if the observations from group $g$ all belong to cluster $c$ and zero otherwise, and
\begin{align*}
\tilde{\bm{e}}' = \tilde{\mathbf{y}}' - \tilde{\mathbf{n}}\odot\tilde{\mathbf{M}}\hat{\bm{\beta}}
\end{align*}
where $\odot$ represents the Hadamard (element-wise) product.


\subsubsection{Between-cluster Compression}
\label{betweenclustercompression}

Panel models that include a time variable are hard to compress according to Section \ref{withinclustercompression}, because user clusters will not have duplicate features. Another approach compresses $\mathbf{M}$ based on identical feature matrices across clusters, rather than single feature vectors. Unlike the previous method, this allows observations from multiple clusters to be mixed into a compressed record. We rewrite the meat matrix as a sum over $G^*$ groups of clusters:
\begin{align*}
    \hat{\mathbf{\Xi}}_\text{NW} &= \sum\limits_{g}^{G^*}\sum\limits_{c \in g}^{n_g} \mathbf{M}_c^\top (\mathbf{y}_c - \mathbf{M}_c \hat{\bm{\beta}}) (\mathbf{y}_c - \mathbf{M}_c \hat{\bm{\beta}})^\top \mathbf{M}_c\\
&= \sum\limits_{g}^{G^*} \left[
        \mathbf{M}_g^\top \left(\textstyle \sum\limits_{c \in g}^{n_g}\mathbf{y}_c\mathbf{y}_c^\top -
        (\sum\limits_{c \in g}^{n_g}\mathbf{y}_c)\hat{\bm{\beta}}^\top\mathbf{M}_g^\top - ((\sum\limits_{c \in g}^{n_g}\mathbf{y}_c)\hat{\bm{\beta}}^\top\mathbf{M}_g^\top)^\top + n_g\mathbf{M}_g\hat{\bm{\beta}}\hat{\bm{\beta}}^\top\mathbf{M}_g^\top\right) \mathbf{M}_g \right],
\end{align*}
and leverage the fact that $\mathbf{M}_c$ is common between $n_g$ clusters in group $g$.
Following the same compression strategy as in Section~\ref{sec:lossless_compression_with_suff_stats},
we create $\tilde{\mathbf{M}}$ by stacking the deduplicated $\mathbf{M}_g$ matrices. The corresponding sufficient statistics are vectors with element $g$ equal to 
$\tilde{\mathbf{y}}'_g = \sum\limits_{c \in g}^{n_g}\mathbf{y}_c$,
$\tilde{\mathbf{y}}_g'' = \sum\limits_{c \in g}^{n_g}\mathbf{y}_c\mathbf{y}_c^\top$ and $\tilde{\mathbf{n}}_g=n_g$. The second sufficient statistic is the sum of outer products, which is a generalization of the previous sum of squares when there is autocorrelation. Its size is quadratic in the number of within-cluster observations.

In our running example of panel data, this compression would result in $G^1 \cdot T$ compressed records, where $G^1$ are the number of unique feature vectors in $\mathbf{M}_1$ alone, since $\mathbf{M}_2$ is perfectly duplicated within clusters. However, it would also introduce $G^1\cdot(T^2 + T)$ additional sufficient statistics due to the outer products and counts of observations.
To be efficient we would thus require
$G^1 \leq \frac{C}{T+2}$.

\subsubsection{Within-cluster Compression on Static Features only}
\label{withinclustercompressionstaticonly}

By leveraging the split between
static and dynamic features, we can make a compression strategy that is applicable for any structure of the feature matrix and always allows us to compress data to $C$ records. Though the previous section shows it is possible to compress to $G^1 T \leq C$ compressed records, this strategy will compress better when each cluster's feature matrix is unique; it is also the only strategy that can guarantee compression to $C$ compressed records while staying robust to a time varying covariate. The strategy is also computationally efficient when interactions are used in the model.

For a given cluster, $c$, we can write $\mathbf{K}_c^1 = \mathbf{M}_c^\top \mathbf{M}_c$, and $\mathbf{K}_c^2 = \mathbf{M}_c^\top \mathbf{y}_c$.
In addition, we will horizontally stack three cluster level variables

\begin{align*}
\begin{bmatrix} \mathbf{K}^1_c \end{bmatrix} 
    &= \begin{bmatrix}\mathbf{K}^1_1 & \mathbf{K}^1_2 & \ldots & \mathbf{K}^1_C \end{bmatrix} \in\mathbb{R}^{p\times pC}, \\
\begin{bmatrix} \mathbf{K}^1_c\hat{\bm{\beta}} \end{bmatrix} 
    &= \begin{bmatrix}\mathbf{K}^1_1\hat{\bm{\beta}} & \mathbf{K}^1_2\hat{\bm{\beta}} & \ldots & \mathbf{K}^1_C \hat{\bm{\beta}} \end{bmatrix} \in\mathbb{R}^{p\times C} \\
    &= \begin{bmatrix} \mathbf{K}^1_c \end{bmatrix}  (\mathbf{I}_{C \times C} \otimes \hat{\bm{\beta}})\text{, and}\\
\begin{bmatrix} \mathbf{K}^2_c \end{bmatrix} 
    &= \begin{bmatrix}\mathbf{K}^2_1 & \mathbf{K}^2_2 & \ldots\end{bmatrix} \in\mathbb{R}^{p\times C},
\end{align*}
where $\otimes$ denotes the Kronecker product, and the brackets around $\begin{bmatrix} \mathbf{X}_c \end{bmatrix}$ represent the horizontal concatenation of $\mathbf{X}_c$ across all values of $c$. This allows us to express $\bm{\hat{\beta}}$, $\mathbf{\Pi}$, and $\hat{\mathbf{\Xi}}_\text{NW}$ in ways that are computationally efficient.
\begin{align*}
    \mathbf{\Pi} &= (\textstyle\sum\limits_c \mathbf{K}_c^1)^{-1} \\
        &= (\begin{bmatrix} \mathbf{K}^1_c \end{bmatrix} \mathbf{I}_{pC \times p})^{-1}, \\
    \hat{\bm{\beta}} &= \mathbf{\Pi} \textstyle\sum\limits_c \mathbf{K}_c^2 \\
        &= \mathbf{\Pi} \begin{bmatrix} \mathbf{K}^2_c \end{bmatrix} \mathbf{1}_C \text{, and}\\
    \hat{\mathbf{\Xi}}_\text{NW} &= \textstyle\sum\limits_c \mathbf{M}_c^\top (\mathbf{y}_c - \mathbf{M}_c \hat{\bm{\beta}}) (\mathbf{y}_c - \mathbf{M}_c \hat{\bm{\beta}})^\top \mathbf{M}_c \\
    &= \textstyle\sum\limits_c (\mathbf{K}_c^2 - \mathbf{K}_c^1 \hat{\bm{\beta}})(\mathbf{K}_c^2 - \mathbf{K}_c^1 \hat{\bm{\beta}})^\top\\
    \begin{split}
        &= \begin{bmatrix} \mathbf{K}^2_c \end{bmatrix} \big(\begin{bmatrix} \mathbf{K}^2_c \end{bmatrix}\big)^{\top} - \begin{bmatrix} \mathbf{K}^1_c\hat{\bm{\beta}} \end{bmatrix} \big(\begin{bmatrix} \mathbf{K}^2_c \end{bmatrix}\big)^{\top} - \\  &\qquad (\mathbf{K}^1\hat{\bm{\beta}}\big(\begin{bmatrix} \mathbf{K}^2_c \end{bmatrix}\big)^{\top})^\top + \begin{bmatrix} \mathbf{K}^1_c\hat{\bm{\beta}} \end{bmatrix} \big(\begin{bmatrix} \mathbf{K}^1_c\hat{\bm{\beta}} \end{bmatrix}\big)^\top.
    \end{split}
\end{align*}

To minimize computation, we reuse a partitioning of the feature matrix into two parts, $\mathbf{M} = \begin{bmatrix}\mathbf{M}_1 & \textbf{M}_2\end{bmatrix}$ where $\mathbf{M}_1$ contains the features that are static within all clusters and $\mathbf{M}_2$ those that change for at least some clusters, for example time. For a cluster, $c$, let $\mathbf{m}_{1,c}^\top$ represent the row vector of the deduplicated rows of $\mathbf{M}_{1,c}$. Then, we can write $\mathbf{M}_{1,c}$ as $\mathbf{1}_{n_c}\mathbf{m}_{1,c}^\top$ where $\mathbf{1}_{n_c}$ is a length $n_c$ column vector of all ones and $n_c$ is the number of records in the cluster.
This structure allows us to reduce $\mathbf{K}_c^1$, $\begin{bmatrix} \mathbf{K}^1_c\hat{\bm{\beta}} \end{bmatrix}$ and $\begin{bmatrix} \mathbf{K}^2_c \end{bmatrix}$ to
\begin{align*}
    \mathbf{K}_c^1 &= \begin{bmatrix}
        \mathbf{M}_{1,c} & \mathbf{M}_{2,c}
        \end{bmatrix}^\top \begin{bmatrix}
        \mathbf{M}_{1,c} & \mathbf{M}_{2,c}
        \end{bmatrix} \\
    &= \begin{bmatrix}
        \mathbf{1}_{n_c}\mathbf{m}_{1,c}^\top & \mathbf{M}_{2,c}
        \end{bmatrix}^\top \begin{bmatrix}
        \mathbf{1}_{n_c}\mathbf{m}_{1,c}^\top & \mathbf{M}_{2,c}
        \end{bmatrix} \\
    &= \begin{bmatrix}
        n_c \mathbf{m}_{1,c} \mathbf{m}_{1,c}^\top & \mathbf{m}_{1,c}\mathbf{1}_{n_c}^\top \mathbf{M}_{2,c} \\
        &
        \mathbf{M}_{2,c}^\top \mathbf{M}_{2,c} \\
        \end{bmatrix}, \\
    %
    \begin{bmatrix} \mathbf{K}^1_c\hat{\bm{\beta}} \end{bmatrix} &= \begin{bmatrix}
        \tilde{\mathbf{M}}_1^\top\text{diag}(\tilde{\mathbf{M}}_1\hat{\bm{\beta}}_1 \odot \tilde{\mathbf{n}})
            + \tilde{\mathbf{M}}_1^\top\text{diag}(\mathbf{W}_C^\top \mathbf{M}_2\hat{\bm{\beta}}_2) \\
        \mathbf{M}_2^\top\mathbf{W}_C\text{diag}(\tilde{\mathbf{M}}_1\hat{\bm{\beta}}_1)
            + \begin{bmatrix} \mathbf{M}_{2,c}^\top \mathbf{M}_{2,c} \end{bmatrix}(\mathbf{I}_C\otimes\hat{\bm{\beta}}_2)
    \end{bmatrix} \text{, and}\\
    \begin{bmatrix} \mathbf{K}^2_c \end{bmatrix} &= \begin{bmatrix}
        \tilde{\mathbf{M}}_1^\top\text{diag}(\tilde{\mathbf{y}}')\\
        \begin{bmatrix}\mathbf{M}_{2,c}^\top \mathbf{y}_c \end{bmatrix}
    \end{bmatrix} \\
    &= \begin{bmatrix}
        \tilde{\mathbf{M}}_1^\top\text{diag}(\tilde{\mathbf{y}}')\\
        \mathbf{M}_2^\top\text{diag}(\mathbf{y})\mathbf{W}_C
    \end{bmatrix},
\end{align*}
where $\mathbf{K}_c^1$ is a symmetric matrix so we omit the lower triangle, $\tilde{\mathbf{M}}_1 \in \mathbb{R}^{C \times p_1}$ are the stacked $\mathbf{m}^\top_{1,c}$ matrices, and $\tilde{\mathbf{y}}'$ is compressed as before. Additionally, $\hat{\bm{\beta}} = \begin{pmatrix}\hat{\bm{\beta}}_1 &  \hat{\bm{\beta}}_2\end{pmatrix}$ with $\hat{\bm{\beta}}_1$ and $\hat{\bm{\beta}}_2$ corresponding to $\mathbf{M}_1$ and $\mathbf{M}_2$ respectively.
The product $\mathbf{\tilde{M}_1 \hat{\bm{\beta}}_1}$ is the contribution to the fitted values of $\mathbf{y}$ from $\tilde{\mathbf{M}}_1$ features.
$\mathbf{W}_C^\top \mathbf{M}_2$ are the column sums
of $\mathbf{M}_2$ per cluster. Based on these reductions the compression method is simple. In addition to $\tilde{\mathbf{M}}_1$ and $\tilde{\mathbf{y}}'$, compute $\mathbf{M}_2^\top\mathbf{W}_C$, $\mathbf{M}_2^\top\text{diag}(\mathbf{y})\mathbf{W}_C$ and $\begin{bmatrix} \mathbf{M}_{2,c}^\top \mathbf{M}_{2,c} \end{bmatrix}$. If $\mathbf{M}_{2}$ is just a time trend, this can be reduced even further.
The matrices needed to recover $\bm{\hat{\beta}}$, $\mathbf{\Pi}$, and $\hat{\mathbf{\Xi}}_\text{NW}$ are sufficient for linear transformations of features in $\mathbf{M}_2$ as long as the parameters are fixed at the cluster level, as is the case for features in $\mathbf{M}_1$. This allows for arguably the most common change to the feature matrix: the addition of interaction terms between features.
Suppose $\mathbf{M}_{3, c}$ is the interaction of $\mathbf{M}_{1, c}$ and $\mathbf{M}_{2, c}$. This gives us
\begin{align*}
    \mathbf{K}_c^1 &= \begin{bmatrix}
        n_c \mathbf{m}_{1, c} \mathbf{m}_{1, c}^\top & \mathbf{m}_{1, c}\mathbf{1}_{n_c}^\top \mathbf{M}_{2, c} & 
        \mathbf{m}_{1, c}\mathbf{1}_{n_c}^\top \mathbf{M}_{3, c} \\
        & \mathbf{M}_{2, c}^\top \mathbf{M}_{2, c} & \mathbf{M}_{2, c}^\top \mathbf{M}_{3, c}\\
        & & \mathbf{M}_{3, c}^\top \mathbf{M}_{3, c}\\
        \end{bmatrix}. \\
    \begin{bmatrix} \mathbf{K}^1_c\hat{\bm{\beta}} \end{bmatrix}
    &= \begin{bmatrix}
        \tilde{\mathbf{M}}_1^\top\text{diag}(\tilde{\mathbf{M}}_1\hat{\bm{\beta}}_1 \odot \tilde{\mathbf{n}})
            + \tilde{\mathbf{M}}_1^\top\text{diag}(\mathbf{W}_C^\top \mathbf{M}_2\hat{\bm{\beta}}_2)
            + \tilde{\mathbf{M}}_1^\top\text{diag}(\mathbf{W}_C^\top \mathbf{M}_3\hat{\bm{\beta}}_3) \\
        \mathbf{M}_2^\top\mathbf{W}_C\text{diag}(\tilde{\mathbf{M}}_1\hat{\bm{\beta}}_1)
            + \begin{bmatrix} \mathbf{M}_{2,c}^\top \mathbf{M}_{2,c} \end{bmatrix}(\mathbf{I}_{C \times C}\otimes\hat{\bm{\beta}}_2)
            + \begin{bmatrix} \mathbf{M}_{2,c}^\top \mathbf{M}_{3,c} \end{bmatrix}(\mathbf{I}_{C \times C}\otimes\hat{\bm{\beta}}_3) \\
        \mathbf{M}_3^\top\mathbf{W}_C\text{diag}(\tilde{\mathbf{M}}_1\hat{\bm{\beta}}_1)
            + \begin{bmatrix} \mathbf{M}_{3,c}^\top \mathbf{M}_{2,c} \end{bmatrix}(\mathbf{I}_{C \times C}\otimes\hat{\bm{\beta}}_2)
            + \begin{bmatrix} \mathbf{M}_{3,c}^\top \mathbf{M}_{3,c} \end{bmatrix}(\mathbf{I}_{C \times C}\otimes\hat{\bm{\beta}}_3)
    \end{bmatrix}. \\
    \begin{bmatrix} \mathbf{K}^2_c \end{bmatrix} &= \begin{bmatrix}
        \tilde{\mathbf{M}}_1^\top\text{diag}(\tilde{\mathbf{y}}')\\
        \mathbf{M}_2^\top\text{diag}(\mathbf{y})\mathbf{W}_C\\
        \mathbf{M}_3^\top\text{diag}(\mathbf{y})\mathbf{W}_C
    \end{bmatrix}.
\end{align*}

When $\mathbf{M}_{2, c}$ is the same for each cluster, as in our motivating example of a balanced panel, we can estimate a model with interactions without constructing the large $\mathbf{M}_3 \in \mathbb{R}^{n \times p_1 p_2}$ matrix. First, we compress $\mathbf{M}_2$ the same way as $\mathbf{M}_1$ to produce $\tilde{\mathbf{M}}_2$. Then, we take advantage of the matrix factorizations $\mathbf{M}_3 = \mathbf{\tilde{M}}_1 \otimes \mathbf{\tilde{M}}_2$, and $\mathbf{W}_C = \mathbf{I}_{C \times C} \otimes \bm{1_T}$. Using properties of the Kronecker product \citep{van2000ubiquitous}, we gain the simplifications

\begin{align*}
\sum\limits_{c} \mathbf{K}_c^1 &= \begin{bmatrix}
        \mathbf{\tilde{M}}_1^\top \text{diag}(\mathbf{\tilde{n}}) \mathbf{\tilde{M}}_1 &
        (\mathbf{1}_{C}^\top \mathbf{\tilde{M}}_1) (\mathbf{\tilde{M}}_2^\top \mathbf{1}_{C}) &
        \mathbf{\tilde{M}}_1^\top
        \bigg(
        \begin{bmatrix}
            \bm{1}_T^\top (\tilde{\mathbf{M}}_{1,c} \otimes \tilde{\mathbf{M}}_2) 
        \end{bmatrix} \bigg)^\top\\ 
    &
    \mathbf{\tilde{M}}_2^\top \text{diag}(C) \mathbf{\tilde{M}}_2 &
    \mathbf{\tilde{M}}_2^\top \big((\mathbf{1}_{C}^\top \mathbf{\tilde{M}}_1) \otimes \mathbf{\tilde{M}}_2\big) \\
    &
    &
    (\mathbf{\tilde{M}}_1^\top \mathbf{\tilde{M}}_1) \otimes (\mathbf{\tilde{M}}_2^\top \mathbf{\tilde{M}}_2)
    \end{bmatrix}. \\
\begin{bmatrix} \mathbf{K}^1_c\hat{\bm{\beta}} \end{bmatrix}
    &= \begin{bmatrix}
    \tilde{\mathbf{M}}_1^\top\text{diag}(
    \tilde{\mathbf{M}}_1\hat{\bm{\beta}}_1 \odot \tilde{\mathbf{n}} +
    \bm{1}_T^\top \tilde{\mathbf{M}}_2 \hat{\bm{\beta}}_2 \bm{1}_C^\top + 
    \bm{1}_T^\top \tilde{\mathbf{M}}_2
            \text{Matrix}(\hat{\bm{\beta}}_3, p_2, p_1) \tilde{\mathbf{M}}_1^\top
    ) \\
            (\bm{1}_C^\top \otimes \tilde{\mathbf{M}}_2^\top \bm{1}_T)
            \text{diag}(\tilde{\mathbf{M}}_1\hat{\bm{\beta}}_1)
            + \bm{1}_C^\top \otimes (\tilde{\mathbf{M}}_2^\top \tilde{\mathbf{M}}_2 \hat{\bm{\beta}}_2)
            + \tilde{\mathbf{M}}_2^\top \tilde{\mathbf{M}}_2 \text{Matrix}(\hat{\bm{\beta}}_3, p_2, p_1) \tilde{\mathbf{M}}_1^\top \\
            \begin{bmatrix}
            \bm{1}_T^\top (\tilde{\mathbf{M}}_{1,c} \otimes \tilde{\mathbf{M}}_2) 
            \end{bmatrix} \text{diag}(\tilde{\mathbf{M}}_1\hat{\bm{\beta}}_1)
            + \begin{bmatrix}
            \text{Vec}(
            \tilde{\mathbf{M}}_2^\top \tilde{\mathbf{M}}_2 \hat{\bm{\beta}}_2 \tilde{\mathbf{M}}_{1, c})
            \end{bmatrix}
            \\ + \begin{bmatrix}
            \text{Vec}(
            \tilde{\mathbf{M}}_2^\top \tilde{\mathbf{M}}_2 \text{Matrix}(\hat{\bm{\beta}}_3, p_2, p_1) \tilde{\mathbf{M}}_{1, c}^\top \tilde{\mathbf{M}}_{1, c})
            \end{bmatrix}
    \end{bmatrix}. \\
    \begin{bmatrix} \mathbf{K}^2_c \end{bmatrix} &= \begin{bmatrix}
        \tilde{\mathbf{M}}_1^\top\text{diag}(\tilde{\mathbf{y}}')\\
        \tilde{\mathbf{M}}_2^\top \text{Matrix}(y, T, C)\\
        \begin{bmatrix}
            \text{Vec}(
            \tilde{\mathbf{M}}_2^\top \mathbf{y}_c \tilde{\mathbf{M}}_{1, c}^\top)
        \end{bmatrix}
    \end{bmatrix}.
\end{align*}
where the operation $\text{Matrix}(\bm{x}, \text{rows}, \text{cols})$ reshapes a vector, $\bm{x}$, into a $\text{rows} \times \text{cols}$ matrix, and $\text{Vec}(\mathbf{X})$ reshapes a matrix, $\mathbf{X}$, into a vector.
In this special case, we also gain optimizations in $\sum\limits_{c} \mathbf{K}_c^1$ used to estimate the parameters.
A derivation is shown in the appendix. In the balanced panel case, the entire model can be estimated by having $\tilde{\mathbf{M}}_1$, $\tilde{\mathbf{M}}_2$, $\tilde{\mathbf{y}}'$, and $\mathbf{y}$.



\subsection{Performance}

Compression reduces data volume, which has a direct consequence on the runtime for fitting linear models. It also has indirect effects, such as making it easier to store all data in memory, improved vectorization, and improved cache hits; all of these also have benefits to performance. Below we summarize the runtime for fitting linear models with different covariances. For homoskedastic and heteroskedastic covariances, the runtime is a function of $G$ compressed records, instead of $n$ individual records, which can lead to orders of magnitude improvement in performance. Similarly, we also see a performance improvement on the order of $T/2$ for clustered covariances, since balanced panel datasets can be compressed from $n_u \cdot T$ records to $n_u$. Most importantly, these linear models can be fit to data at interactive speeds.

\begin{figure}[h]
\centering
\caption{Performance Benchmark}
\label{fig:performance_benchmark}
\begin{subfigure}{.325\textwidth}
  \centering
  \includegraphics[width=.95\linewidth]{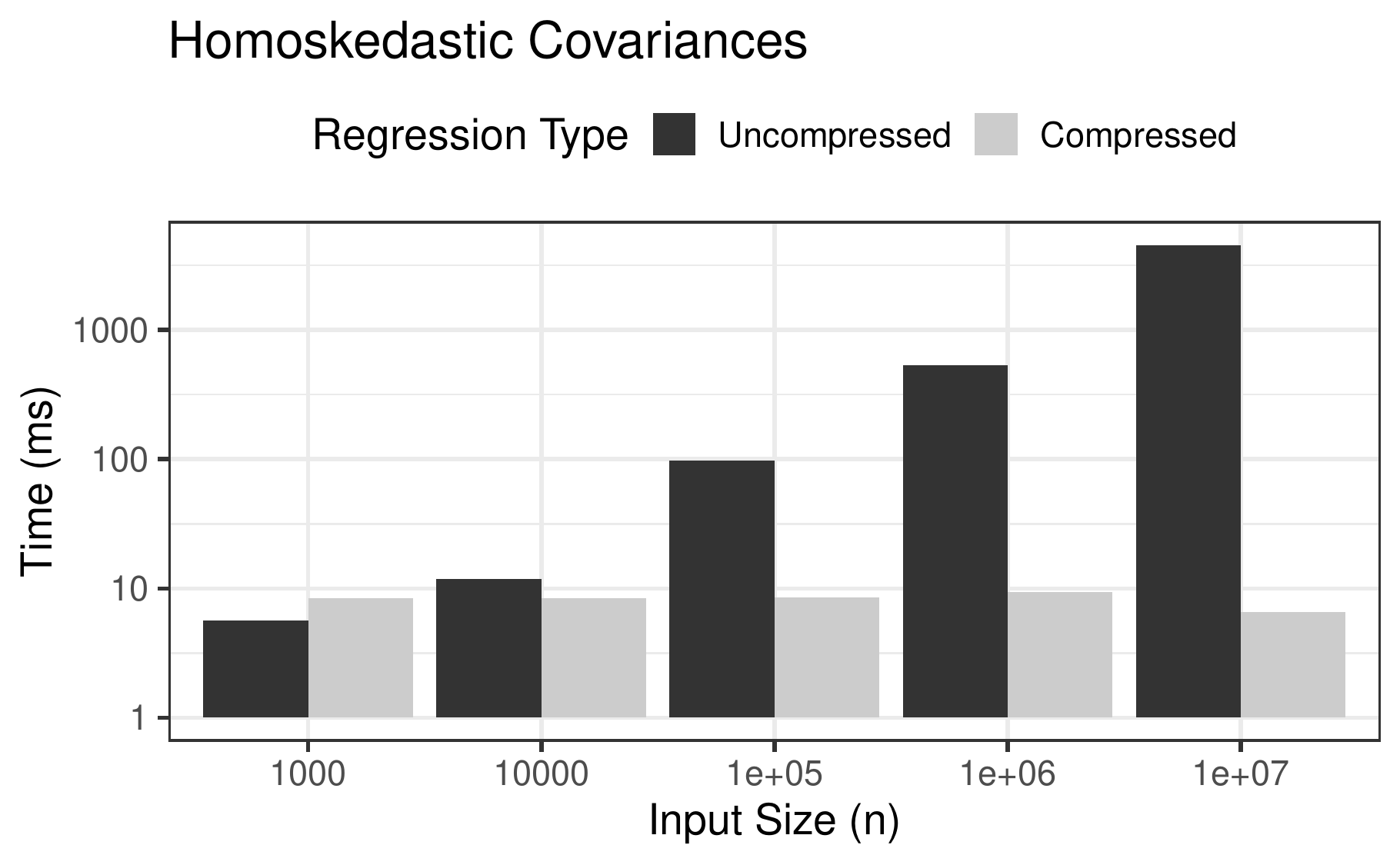}  
  \label{fig:sub-benchmarks_homoskedastic}
\end{subfigure}
\begin{subfigure}{.325\textwidth}
  \centering
  \includegraphics[width=.95\linewidth]{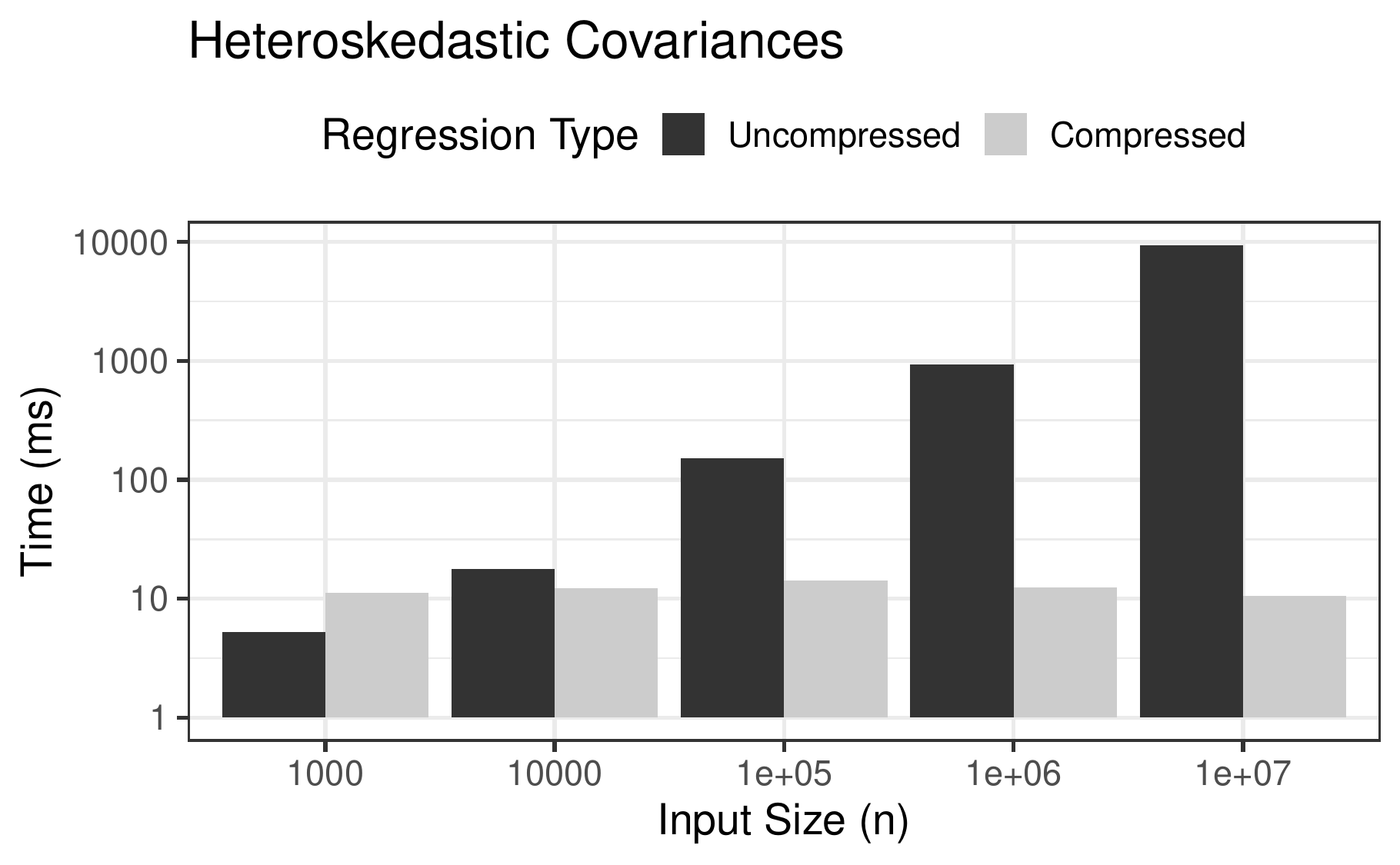}  
  \label{fig:sub-benchmarks_heteroskedastic}
\end{subfigure}
\begin{subfigure}{.325\textwidth}
  \centering
  \includegraphics[width=.95\linewidth]{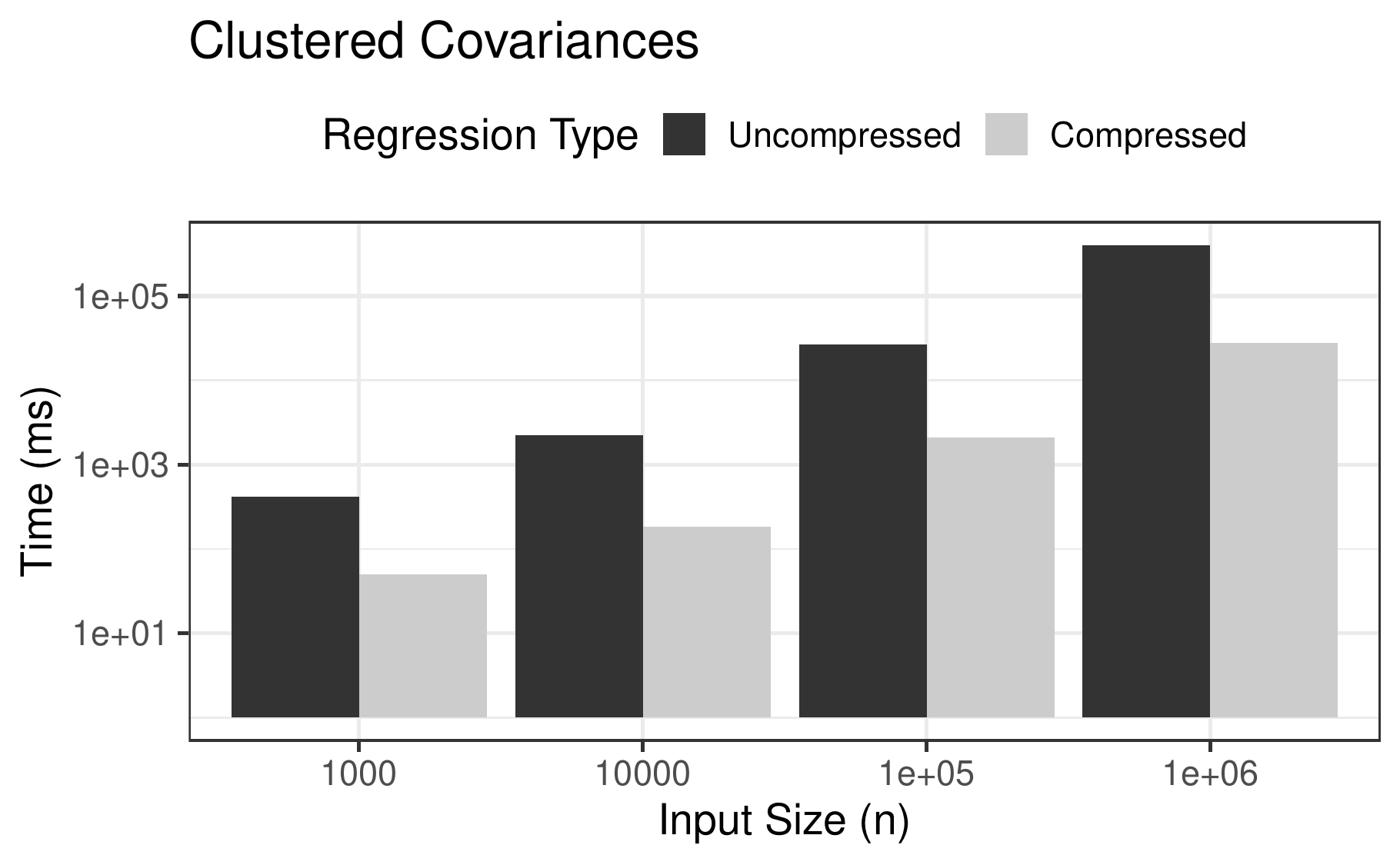}
  \label{fig:sub-benchmarks_clustering}
\end{subfigure}
\end{figure}

\section{Features with High Cardinality}

Binning or rounding features can make compression practical when the number of unique feature vectors in $\mathbf{M}$ is large.
While changing the features in the model will change the regression coefficients, $\bm{\hat{\beta}}$, predictions for $\hat{\mathbf{y}}$, RSS, and $\mathbb{V}(\bm{\hat{\beta}})$, we can show that under practical conditions the estimator for the treatment effects is consistent for the true treatment effect, and endogeneity through measurement error \citep{woolridgeMeasurementError} does not occur. 
First, we consider a new partitioning of $\mathbf{M}$ into two types of features: $\mathbf{A}$ and $\mathbf{X}$. Let $\mathbf{A}$ represent the treatment variable, which only has a few unique values, and $\mathbf{X}$ represent additional pretreatment covariates, which have many unique values. 
If $\mathbf{X}$ is binned, it continues to be an exogeneous pretreatment variable, and therefore estimators for the treatment effect are still consistent. Simultaneously, the feature matrix has fewer unique feature vectors, yielding a better compression rate.

In addition, binning is a feature transform that is broadly useful for engineering systems. 
Suppose the true form of the data generating process is $\mathbf{y} = \alpha + f(\mathbf{A})\bm{\beta}_1 + g(\mathbf{X})\bm{\beta}_2 + h(f(\mathbf{A}), g(\mathbf{X}))\bm{\beta}_3 + \bm{\varepsilon}$, where $f$, $g$, and $h$ are unknown, nonlinear functions on the features. Learning these forms can reduce variance on the treatment effect. Engineering systems that are general may not be able to leverage context or domain knowledge to utilize these forms, however it can achieve a general, nonlinear transformation on the features by binning $\mathbf{X}$, for example through decile binning, and regressing on subsequent dummy variables. This allows the system an improvement in the compression rate, while also gaining nonlinear features. Furthermore, estimating heterogeneous effects can be sensitive and biased depending on the form of of the model, but \citet{ATHEY201773} argues that interacting dummy variables is the only way to have an unbiased estimate of a heterogeneous effect.

\section{Extensions}

\subsection{Multiple Outcome Variables}

Consider the multiple outcome variables case where $o > 1$. We observe $\begin{pmatrix}\mathbf{y}_i^\top & \mathbf{m}_i^\top\end{pmatrix}$ where $\mathbf{y}_i^\top$ is a row vector of outcomes for each observation. We can compress as above, keeping track of sufficient statistics per outcome variable. Finally, $\hat{\bm{\beta}} \in \mathbb{R}^{p \times o}$ can be estimated simultaneously for multiple outcome variables by horizontally concatenating $\tilde{\mathbf{y}}'$ from each outcome and estimating the model
$\frac{\tilde{\mathbf{y}}'}{\tilde{\mathbf{n}}} = \tilde{\mathbf{M}} \hat{\bm{\beta}} + \bm{\varepsilon}.$

\subsection{Estimating OLS with Compression and Other Weights}

The previous sections of this paper discuss how to cast an unweighted OLS problem into a weighted OLS problem with compressed data. We now show how to adapt the compression and estimation techniques when the original problem also contains weights. Other than frequency weights, a regression problem may have analytic weights, probability weights, or importance weights. 
Without loss of generality, we denote these types of weights as $\mathbf{w}$.

Suppose we collect data in the form $\begin{pmatrix}
y_i &  
\mathbf{m}_i^\top &
w_i
\end{pmatrix}$. We wish to compress the data and combine the group sizes with $\mathbf{w}$ to learn a weighted linear model. Despite adding more information to each observation, we can deduplicate according to $\mathbf{m}_i^\top$ alone, just as before; the presence of a continuous value for $w_i$ does not affect the compression rate. First, we define functions for weighted conditionally sufficient statistics $T(\mathbf{y}, \mathbf{w} | \mathbf{m^*}) = \{ \sum\limits_{i | \mathbf{m}_i = \mathbf{m}^*} y_i w_i,\;\sum\limits_{i | \mathbf{m}_i = \mathbf{m}^*} y_i^2 w_i,\;\sum\limits_{i | \mathbf{m}_i = \mathbf{m}^*} w_i \}$. Then, $\tilde{\mathbf{y}}'(\mathbf{w})$, $\tilde{\mathbf{y}}''(\mathbf{w})$ and $\tilde{\mathbf{w}}(\mathbf{w})$ output three column vectors of sufficient statistics for $\mathbf{\tilde{m}}_1^\top \ldots \mathbf{\tilde{m}}_G^\top$ just as before. Unweighted sufficient statistics can be thought of as output of these functions with $\mathbf{w} = \mathbf{1}_n$. The parameter estimates are

\begin{align*}
    \hat{\bm{\beta}} = (\tilde{\mathbf{M}}^\top \text{diag}(\tilde{\mathbf{w}}) \tilde{\mathbf{M}})^{-1} (\tilde{\mathbf{M}}^\top \tilde{\mathbf{y}}'(w)).
\end{align*}
For homoskedastic covariances, we have weighted residual sum of squares that yield

\begin{align*}
    WSS &= \sum\limits_{g = 1}^G \hat{\tilde{y}}_{g}^2 \tilde{{w}}_g - 
    2 \hat{\tilde{y}}_{g} \tilde{{y}}'_g(w_g) + \tilde{{y}}''_g(w_g) = \sum\limits_{g = 1}^G \widetilde{WSS}_g, \\
    \hat{\sigma}^2 &= \frac{WSS}{n - p} \text{, and}\\
    \mathbb{V}(\hat{\beta}) &= (\tilde{\mathbf{M}}^\top \text{diag}(\tilde{\mathbf{w}}) \tilde{\mathbf{M}})^{-1} \hat{\sigma}^2.
\end{align*}
With the exception when $\mathbf{w}$ are frequency weights, $\hat{\sigma}^2$ should be $\frac{WSS}{\sum_i w_i - p}$. The bread and meat matrices for heteroskedasticity-consistent covariances are

\begin{align*}
    \mathbf{\Pi} &= (\tilde{\mathbf{M}}^\top \text{diag}(\tilde{\mathbf{w}}) \tilde{\mathbf{M}})^{-1}, \\
    \widetilde{\bm{WSS}} &= 
    \hat{\tilde{\mathbf{y}}}^2 \odot \tilde{ \mathbf{w}}(\mathbf{w}^2) -
      2 \cdot \hat{\tilde{\mathbf{y}}} \odot \tilde{\mathbf{y}}'(\mathbf{w}^2) +
      \tilde{\mathbf{y}}''(\mathbf{w}^2) \text{, and}\\
    \hat{\mathbf{\Xi}}_\text{EHW} &= \tilde{\mathbf{M}}^\top \text{diag}(\widetilde{\bm{WSS}}) \tilde{\mathbf{M}}.
\end{align*}

\subsection{Compression in Logistic Regression}

Compression via sufficient statistics is not only applicable to OLS, we now show how it is applied to logistic regression.
In this scenario, we record 
$\begin{pmatrix}
y_i &  
\mathbf{m}_i^\top
\end{pmatrix}$ where $y_i$ is either 0 or 1. We deduplicate according to $\mathbf{m}_i^\top$ as before, then aggregate $T(\mathbf{y} | \mathbf{m}^*) = \{ \sum\limits_{i | \mathbf{m}_i = \mathbf{m}^*} y_i, \sum\limits_{i | \mathbf{m}_i = \mathbf{m}^*} 1 \}$, omitting the sum of squares of $y$ since it is not a sufficient statistic for the binomial distribution.

Logistic regression estimates the linear model
$$\log{\frac{\mathbf{p}}{1-\mathbf{p}}} = \mathbf{M}\bm{\beta} + \bm{\varepsilon},$$
by maximizing the log likelihood function

\begin{align*}
    l(\bm{\beta}) &= \sum\limits_{i = 1}^n y_i \log(s(\mathbf{m}_i^\top \bm{\beta})) + (1 - y_i) \log(1 - s(\mathbf{m}_i^\top \bm{\beta})) \text{, with}\\
    s(z) &= \frac{1}{1 + e^{-z}}.
\end{align*}
Given the sufficient statistics, this is simply rewritten as 

\begin{align*}
    l(\bm{\beta}) &= \sum\limits_{g = 1}^G \tilde{y}_g' \log(s(\mathbf{\tilde{m}}_g^\top \bm{\beta})) + (\tilde{n}_g - \tilde{y}_g') \log(1 - s(\mathbf{\tilde{m}}_g^\top \bm{\beta})).
\end{align*}
This allows all solvers to iterate on compressed records. The covariance matrix of the logistic regression parameters \citep{hosmer2013applied} is

$$\mathbb{V}(\hat{\bm{\beta}}) = \tilde{\mathbf{M}}^\top \mathbf{W}^{\text{LR}} \tilde{\mathbf{M}},$$
where $\mathbf{W}^{\text{LR}}$ is a diagonal matrix with g-th diagonal entry equal to:
$$s^{-1} (\mathbf{\tilde{m}}_g^\top \bm{\beta}) (1 - s^{-1} (\mathbf{\tilde{m}}_g^\top \bm{\beta})) \tilde{n}_g.$$

\section{Conclusion}
Data compression is particularly important in managing engineering systems that analyze data - it decreases memory consumption, network latency, and makes statistical modeling computationally performant. At the same time, modeling software that is more efficient improves research productivity.
We have shown that grouping features and aggregating sufficient statistics, for example in a database or on a data frame, can compress the volume of data needed to estimate linear models without loss. As opposed to traditional approaches like frequency weighting, this compression only relies on duplication of the features used in the model, not of the outcomes, making it versatile. In addition, compression can be achieved with high cardinality features by binning or rounding, which has worthwhile properties to estimating heterogeneous treatment effects as well. Finally, we have shown how the compression strategy is compatible with different types of weights and that it readily applies to logistic regression.

Compression drives productivity improvements across research and engineering. 
The synergies between these enable researchers and engineers to explore data, train models efficiently and locally, and still use the same single-machine code in large scale engineering systems. Ultimately, this aligns offline development and online deployment, removing barriers to integrating linear models into large engineering systems, such as online experimentation platforms.

\bibliography{references}

\appendix
\section{Balanced Panel Compression}

This section lists important reductions that are leveraged in order to derive the compression strategy for balanced panels in Section \ref{withinclustercompressionstaticonly}.

\begin{enumerate}
    \item $\mathbf{M}_2 = \bm{1}_C \otimes \tilde{\mathbf{M}}_2$, $\begin{bmatrix}\mathbf{M}_{2,c}^\top \mathbf{M}_{2,c}\end{bmatrix} = \bm{1}_C^\top \otimes \tilde{\mathbf{M}}_2^\top \tilde{\mathbf{M}}_2$, and   $\mathbf{M}_3^\top \mathbf{M}_3 = (\tilde{\mathbf{M}}_1^\top \tilde{\mathbf{M}}_1) \otimes (\tilde{\mathbf{M}}_2^\top \tilde{\mathbf{M}}_2)$.
    \item $\begin{bmatrix}{\mathbf{M}_{2,c}^\top\mathbf{M}_{3,c}}\end{bmatrix} (\mathbf{I}_{C \times C} \otimes \hat{\bm{\beta}}_3) = \begin{bmatrix}
    \tilde{\mathbf{M}}_{2}^\top (\tilde{\mathbf{M}}_{1, 1} \otimes \tilde{\mathbf{M}}_2) \hat{\bm{\beta}}_3 & \cdots & \tilde{\mathbf{M}}_2^\top (\tilde{\mathbf{M}}_{1, C} \otimes \tilde{\mathbf{M}}_2) \hat{\bm{\beta}}_3
    \end{bmatrix}$. When each row of $\tilde{\mathbf{M}}_1$ is a cluster this reduces to $(\tilde{\mathbf{M}}_2^\top \tilde{\mathbf{M}}_2) \text{Matrix}(\hat{\bm{\beta}}_3, p_2, p_1) \tilde{\mathbf{M}}_1$.
    \item $\begin{bmatrix}{\mathbf{M}_{3,c}^\top\mathbf{M}_{2,c}}\end{bmatrix}  (\mathbf{I}_{C \times C} \otimes \hat{\bm{\beta}}_2) = 
    \begin{bmatrix}
    (\tilde{\mathbf{M}}_{1, 1}^\top \otimes \tilde{\mathbf{M}}_2^\top) \tilde{\mathbf{M}}_2 \hat{\bm{\beta}}_2 & \cdots & (\tilde{\mathbf{M}}_{1, C}^\top \otimes \tilde{\mathbf{M}}_2^\top) \tilde{\mathbf{M}}_2 \hat{\bm{\beta}}_2
    \end{bmatrix}$. This is a horizontally stacked matrix with $C$ components. Each component, $c$, can be reduced to $\text{Vec}(\tilde{\mathbf{M}}_2^\top \tilde{\mathbf{M}}_2 \hat{\bm{\beta}}_2 \tilde{\mathbf{M}}_{1, c})$. Likewise, $\begin{bmatrix}{\mathbf{M}_{3,c}^\top\mathbf{M}_{3,c}}\end{bmatrix} (\mathbf{I}_{C \times C} \otimes \hat{\bm{\beta}}_3)$ is also a horizontally stacked matrix, with component $c$ equal to $\text{Vec}(\tilde{\mathbf{M}}_2^\top \tilde{\mathbf{M}}_2 \text{Matrix}(\hat{\bm{\beta}}_3, p_2, p_1) \tilde{\mathbf{M}}_{1, c}^\top \tilde{\mathbf{M}}_{1, c})$.
\end{enumerate}

We also gain structure in $\mathbf{K}^2$ in a balanced panel.
\begin{enumerate}
    \item $\mathbf{W}_C \mathbf{W}_C^\top \in \mathbb{R}^{CT \times CT}$ is a block diagonal matrix $\begin{bmatrix}
    \bm{1}_{T \times T} & \ldots & 0\\
    \vdots & \ddots & \vdots \\
    0 & \ldots & \bm{1}_{T \times T}
    \end{bmatrix}$.
    \item $\text{diag}(\mathbf{y}) \mathbf{W_C} \in \mathbb{R}^{CT \times C}$ is a block band matrix with structure $\begin{bmatrix}
    y_1 & \mathbf{0}_T & \ldots\\
    \vdots & y_{T+1} & \ldots\\
    y_T & \vdots & \ldots\\
    \mathbf{0}_{T} & y_{2T} & \ldots \\
    \mathbf{0}_{(C-2)T} & \mathbf{0}_{(C-2)T} & \ldots
    \end{bmatrix}$.
    \item $\mathbf{M}_2^\top \text{diag}(\mathbf{y}) \mathbf{W}_C = (\bm{1}_C^\top \otimes \tilde{\mathbf{M}}_2^\top) \text{diag}(\mathbf{y}) \mathbf{W}_C \in \mathbb{R}^{p_2 \times C}$. Because $\text{diag}(\mathbf{y}) \mathbf{W_C}$ is a block band matrix where each column has exactly $T$ nonzeros, the product reduces to $\tilde{\mathbf{M}}_2^\top \text{Matrix}(y, T, C)$. Using this insight, $\mathbf{M}_3^\top \text{diag}(\mathbf{y}) \mathbf{W}_C \in \mathbb{R}^{p_3 \times C}$ is a matrix whose j-th column is the outer product of the j-th column of $\mathbf{M}_2^\top \text{diag}(\mathbf{y}) \mathbf{W}_C$ and the j-th row of $\tilde{\mathbf{M}}_1$.
\end{enumerate}

\end{document}